\documentclass[10pt,a4paper,final]{report}
\usepackage[utf8]{inputenc}
\usepackage[numbers]{natbib}
\usepackage{amsmath}
\usepackage{amsfonts}
\usepackage{amssymb}
\usepackage{cancel}
\usepackage{graphicx}
\usepackage{url}
\graphicspath{ {fig/} }
\usepackage{graphicx}

\title{Variational Autoencoders}
\author{Vasanth Kalingeri}
\date{}
\AtBeginDocument{}
\begin{document}
\begin{titlepage}
    \begin{center}
        \vspace*{1cm}
        
        \textbf{\LARGE Latent Variable Modelling \\Using \\Variational Autoencoders: A Survey}
        
        \vspace{1.5cm}
        
        \textbf{Author:} Vasanth Kalingeri\\
    \end{center}
\end{titlepage}
\tableofcontents
\chapter{Who should read this?}
A reader who has a basic idea of machine learning but wants to learn about general themes in machine learning research can benefit from the report. The report explains central ideas on learning probability distributions, what people did to make this tractable and goes into details around how deep learning is currently applied.

The report also serves a gentle introduction for someone looking to contribute to this sub-field.

\chapter{Introduction}
A probability distribution allows practitioners to uncover hidden structure in the data and build models to solve supervised learning problems using limited data. The focus of this report is on Variational autoencoders, a method to learn the probability distribution of large complex datasets. The report provides a theoretical understanding of variational autoencoders and consolidates the current research in the field.
\\

The report is divided into multiple chapters, the first chapter introduces the problem, describes variational autoencoders and identifies key research directions in the field. Chapters 2, 3, 4 and 5 dive into the details of each of the key research areas. Chapter 6 concludes the report and suggests directions for future work. 
\\

This chapter is divided into 5 sections and it provides the foundation for understanding research on variational autoencoders. The first section formally introduces the problem and the concept of latent variable models. Sections 1.2 and 1.3 introduce expectation maximization and variational inference which are methods to learn latent variable models of small datasets under strong assumptions. Section 1.4 introduces variational autoencoders as a faster variant of variational inference that scales to larger datasets. Section 1.5 discusses the problems with variational autoencoders and identifies the key research directions. 
\newpage

\section{The problem} \label{the_problem}
Consider X to be a D dimensional random variable representing the data where each data sample is represented by D dimensional random value x. Here D refers to the dimensionality of the data. For example, X can represent the set of all nxn images, where each image is represented as x, a vector of size n$^2$. We are interested in learning the probability distribution of X, represented as P(X). 

Maximum likelihood estimation is the standard approach to learning unknown probability distributions over data samples. The method proceeds by assuming a probability distribution with unknown parameters and finds the setting of these parameters that maximizes the total log probability of the data. Formally, it assumes some probability distribution over the data with parameters $\theta$, P$_{\theta}$(X) and finds $\theta$ by maximizing $\sum_x$ log P$_{\theta}$(x) $\forall$ x $\epsilon$ data. This method works well as long we choose the right parametric distribution P$_{\theta}$(X), a distribution that can accurately model the data. 

For complex data sets like images, audio and text, choosing the right distribution is hard, to overcome this, we introduce latent variable models. The key idea of latent variable models is to learn extra information about the data so that the distribution of the data conditioned on this information is simpler and easier to model. For example, the distribution over all set of images is largely simplified given the information describing the image contents. In latent variable models, we are interested in finding such information about the data. 

Latent information can be represented using random variables called latent variables. Let Z denote a H dimensional latent variable and z its random value. Finding the right z for a given x is a hard problem, we accomplish this by learning the distribution over the latent states. This breaks up the original problem of learning the distribution P(X) into learning the conditional distribution over the data P(X$|$Z) and learning the distribution over the latent variables P(Z). Following the maximum likelihood principle, we can assume $\theta$ and $\phi$ as unknown parameters of the distributions P$_{\theta}$(X$|$Z) and P$_{\phi}$(Z) respectively and maximize the marginal log probability over all data samples. The marginal log probability for a particular data sample x is given as:
\begin{equation*}
log P(x) = log \sum_{z} P_{\theta}(x | z)P_{\phi}(z)
\end{equation*}
Here $\sum_{z}$ denotes the summation over the latent states. 

(Note: For convenience, the summation sign used in the report is used to represent summation over discrete quantities and integrals over continuous quantities.)
\\

In this view, latent variable models can be seen as a divide and conquer technique, where a complex distribution P(X) is broken down as a summation over simple distributions, $P_{\theta}(x | z)$ and $P_{\phi}(z)$. 

The presence of the summation prevents direct optimization since log P(X) cannot be easily expressed as a function of the parameters $\theta$ and $\phi$. To allow optimization, approximations of log P(x) are used, the next section derives one such approximation and describes maximum likelihood estimation using the expectation maximization algorithm. 

\section{Expectation Maximization}
Expectation maximization(EM) is an algorithm to perform maximum likelihood estimation in latent variable models, it works by optimizing an objective that is an exact approximation of the marginal log probability of the data.

The approximation is obtained by assuming a distribution q(Z) over the latent states and deriving a lower bound on the marginal log probability. The approximation is made exact by choosing q(Z) as the posterior P(Z$|$X).

The derivation follows from the property that the summation over the latent states can be replaced by an expectation under any distribution q(Z) as long as q(Z) $>$ 0. This can be seen as follows:
\begin{equation*}
log P(x) = log \sum_{z} P(x, z) = log E_{q(Z)} \left[\frac{P(x, z)}{q(z)}\right]
\end{equation*}

Using Jenson's inequality, the expectation of the log can be used to lower bound the log of the expectation, thereby lower bounding the marginal log probability. This lower bound allows optimization and is called the evidence lower bound(ELBO). Applying Jenson's inequality to the above equation, we get:
\begin{equation}
\underbrace{E_{q(Z)} \left[ log \frac{P(x, z)}{q(z)}\right]}_\textrm{Evidence Lower Bound (ELBO)} \leq log P(x)
\end{equation}

During optimization, we are interested in maximizing the log probability over all the samples $\sum_x log P(x)$. When considering all samples, ELBO is expressed as:
\begin{equation}
E_{q(Z)} \left[ \sum_x log \frac{P(x, z)}{q(z)}\right] \leq log P(x)
\end{equation}
Any method that works by maximizing ELBO by computing the expectation in closed form cannot be extended to large datasets since this summation over all data samples becomes intractable. To avoid excessive clutter, all following equations in the report are expressed with respect to a single sample.  

The bound is exact when we substitute P(Z$|$X) in place of q(Z), this gives us the objective:
\begin{equation*}
\textrm{Objective} = E_{P(Z|X)} \left[ log \frac{P(x, z)}{P(z|x)}\right] = log P(x)
\end{equation*}

To compute the objective, parametric distributions  $P_{\theta}(x | z)$ and $P_{\phi}(z)$ can be assumed and P(x, z) and P(z$|$x) can be obtained using the following equations:
\begin{equation*}
P(x, z) = P_{\theta}(x | z)P_{\phi}(z)
\end{equation*}
\begin{equation*}
P(z | x) = \frac{P_{\theta}(x | z)P_{\phi}(z)}{\sum_{z} P_{\theta}(x | z)P_{\phi}(z)}
\end{equation*}

The expectation under the posterior is expanded or computed in closed form to obtain an equation in terms of $\theta$, $\phi$ and the posterior. Before the start of optimization, the posterior for each sample is computed using the initial values of $\theta$ and $\phi$, the optimal values of $\theta$ and $\phi$ are then determined by maximizing the objective. $\theta$, $\phi$ obtained after optimization are used to recompute the posterior to repeat the process.

EM requires us to be able to compute the posterior and the expectation under the posterior in closed form. This becomes intractable when $\sum_{z} P_{\theta}(x | z)P_{\phi}(z)$ is intractable, so the number of latent states and the distributions P$_{\theta}(x | z)$ and P$_{\phi}(z)$ have to be carefully chosen to keep optimization tractable. For this reason, EM cannot model complex data and since it computes the expectation in closed form it also cannot deal with large datasets. The next section introduces variational inference that partly lifts these restrictions at the expense of an exact approximation. 
\newpage

\section{Variational Inference}
EM introduces the posterior into its objective to make optimization equivalent to maximum likelihood estimation, however, the inclusion of the posterior sometimes makes EM intractable. To avoid intractability, Variational inference(VI) chooses to maximize ELBO without substituting the posterior. The hope is that since maximizing ELBO makes it a better approximation of the marginal log probability, eventually, when the bound is exact, variational inference will become equivalent to maximum likelihood estimation. 

In EM, q(Z) is the posterior, so it is computed using P$_{\theta}(X|Z)$ and P$_{\phi}(Z)$ as discussed in the previous section. In VI, q(Z) is unknown, so a parametric distribution for q, q$_{\phi}$(Z) is assumed and this distribution q$_{\phi}$(Z) along with P$_{\theta}(X|Z)$ is learnt by maximizing ELBO. The practitioner chooses the prior distribution P(Z) based on the problem. 

Maximizing ELBO makes the bound tighter until it becomes exact, we know that q(Z) is equal to the posterior P(Z$|$X) when the bound is exact, in this view, q$_{\phi}$(Z) can be seen as a distribution that approaches the posterior as ELBO is maximized. The following reformulation of ELBO makes this relationship explicit:
\begin{equation}\label{kl_posterior}
E_{q(Z)} \left[ log \frac{P(x)P(z | x)}{q(z)}\right] = logP(x) - \underbrace{E_{q(Z)} \left[ log q(z) - log P(z|x) \right]}_{KL(q(Z) || P(Z | X))}
\end{equation}
Here KL denotes the Kullback–Leibler divergence between the distributions. 

Thus, as ELBO is maximized, the KL divergence between q$_{\phi}$(Z) and the posterior P(Z $|$ X) keeps decreasing until the bound is exact where it becomes zero. Thus at any instant during optimization, q$_{\phi}$(Z) can be seen as an approximation of the posterior, for this reason it is commonly referred to as the approximate posterior and is sometimes denoted as q$_{\phi}$(Z $|$ X). 

The above reformulation of ELBO is a function of the posterior, since it is intractable, the function cannot be optimized. To allow optimization, ELBO has to be expressed solely in terms of the prior P(Z), the approximate posterior q$_{\phi}$(Z $|$ X) and the conditional distribution of the data P$_{\theta}(X|Z)$ as:
\begin{equation}
ELBO(\theta, \phi) = E_{q_{\phi}(Z)} \left[ log P_{\theta}(x | z)\right] - KL(q_{\phi}(Z) || P(Z))
\end{equation}

The gradient of ELBO with respect to $\phi$ is a function of $\theta$ and vice versa, due to this interdependence, simultaneous optimization of ELBO with respect to $\theta$ and $\phi$ is not possible. Hence, optimization proceeds by repeatedly alternating between maximizing ELBO with respect to $\phi$ treating $\theta$ as a constant, followed by maximizing ELBO with respect to $\theta$ treating $\phi$ as a constant. In order to keep optimization tractable, the parametric distributions and the prior are chosen such that the expectation E$_{q_{\phi}(Z)}[.]$ and the KL divergence can be computed in closed form. 

Computing the distribution over the latent variables given the data sample is termed as inference. VI largely differs from EM in terms of the time taken for inference. In EM, inference is performed by computing the posterior P(z$|$x) $\forall$ z $\epsilon$ Z, since the equation of the posterior is known the time taken to compute it is deterministic. In VI, inference involves finding the approximate posterior q$_{\phi}$(Z $|$ X), which in turn involves finding $\phi$ that maximizes ELBO for the given sample, since optimization time is non-deterministic, inference may become time consuming. To keep inference fast, the parametric distributions and the prior are chosen such that the value of $\phi$ that maximizes ELBO can be computed analytically. 

VI can model data that EM cannot since it can handle cases where the posterior is intractable, however, the set of candidate parametric distributions that can be used is still limited by the need to keep inference fast and optimization tractable. For this reason, VI struggles to model complex data, since it computes expectation in closed form like EM, it cannot deal with large datasets as well. In the next section we discuss Variational autoencoders which overcome these limitations.
\newpage

\section{Variational Autoencoders}
EM approximated marginal log likelihood to make optimization tractable, VI approximated the posterior to make EM tractable, similarly, Variational Autoencoders(VAE) approximate the bottlenecks in VI, the intractable expectation terms in ELBO and the optimization required for inference. Approximating intractable expectation terms scales VI to larger complex datasets and approximating optimization makes inference fast and removes restrictions imposed on the complexity of the approximate posterior.  

The expectation is approximated using the Monte-Carlo technique, according to this method, the expected value of a function under a distribution can be approximated by averaging the function values computed using different samples from the distribution as:
\begin{equation*}
E_{q(Z)}[f(z)] \approx \frac{1}{S} \sum_{i=1}^S f(z_i)
\end{equation*}
Here S denotes the number of samples, and z$_i$ denotes the i$^{th}$ sample obtained from q(Z). The intractable expectations in ELBO, $E_{q_{\phi}(Z)} \left[ log P_{\theta}(x | z)\right]$ and sometimes the KL divergence, are replaced with this approximation.

Optimization for inference involves finding the parameters $\phi$ for a given x, so a function that approximates it should map the input x to the parameters $\phi$. Variational autoencoders use a neural network to learn this mapping, this network outputs the parameters $\phi$ of the approximate posterior q$_{\phi}$(Z$|$X) for a given input x, thus inference in VAEs just involves computing the forward pass through this network.   

Latent information decides the distribution of the data conditioned on it, for example, the distributions of a set of images conditioned on its content vary based on the content, in other words, the parameters $\theta$ of P$_{\theta}(X|Z)$ depend on the value taken by Z. VAEs make this relationship explicit by learning a parametric function that maps the latent state z to the parameter $\theta$ of P$_{\theta}(X|Z)$. This mapping is also learnt using a neural network.

Thus a variational autoencoder uses two neural networks, one to learn the parameters $\phi$ based on x and another to learn $\theta$ based on z. The latent state z is obtained by sampling from the approximate posterior q$_{\phi}(Z|X)$. This setup of having two neural networks work together to model the input is called as an autoencoder, since the networks are trained by maximizing ELBO, the setup performs variational inference, hence the name variational autoencoder. Following this terminology, the network that outputs the parameters $\phi$ is called the encoder and the other network is called the decoder.

Assuming that the distribution of the prior and the approximate posterior are chosen such that the KL divergence between them is tractable, we can express ELBO for a single sample x as:
\begin{equation*}
ELBO(\theta, \phi) = \underbrace{\frac{1}{S} \sum_{i=1}^S \left( log P_{\theta}(x|z_i)\right)}_{E_{q_{\phi}(Z)} \left[ log P_{\theta}(x | z)\right]} - KL(q_{\phi}(Z)||P(Z))
\end{equation*}
Here S denotes the number of samples where each sample z$_i$ is obtained from the approximate posterior. To make computation faster, only a single sample is used to approximate the expectation term, obtaining the objective:
\begin{equation*}
ELBO(\theta, \phi) = log P_{\theta}(x|z) - KL(q_{\phi}(Z)||P(Z))
\end{equation*}

Considering all the data samples x $\epsilon$ X, the total objective maximized by the networks is given as:
\begin{equation}
ELBO(\theta, \phi) = \underbrace{\sum_x log P_{\theta}(x|z)}_{E_{q_{\phi}(Z)} \left[ \sum_x log P_{\theta}(x | z)\right]} - \sum_x KL(q_{\phi}(Z)||P(Z))
\end{equation}
Since the gradient of the above objective can be approximated using minibatches, stochastic gradient descent \cite{robbins1951stochastic} can be used, this is the reason Variational Autoencoders scale to large datasets. 

The encoder outputs the parameters $\phi$ while the decoder takes the latent state z as input, so the two networks have to be connected to each other using the sampling function that takes $\phi$ as input and samples z from $q_{\phi}(Z)$. The gradient of $log P_{\theta}(x|z)$ is computed using only the output of the decoder $\theta$ and the sample x, in order for this loss to be backpropagated from the decoder into the encoder, the sampling function has to be differentiable with respect to its inputs $\phi$. The major contributions of work that introduced variational autoencoders \cite{kingma2013auto}, \cite{rezende2014stochastic}, \cite{titsias2014doubly} was in recognizing parametric distributions $q_{\phi}(Z)$ with a differentiable sampling function, which they call the reparameterization function. 

\begin{figure}[h]
\center
\includegraphics[scale=0.3]{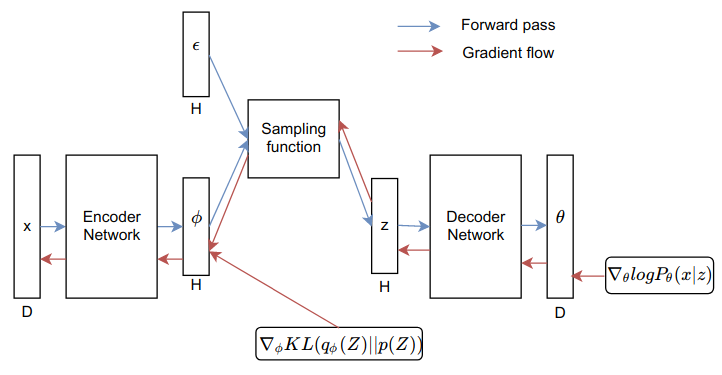}
\caption{The architecture of VAE. Vectors are shown as long rectangular blocks with their dimensions mentioned below it. Gradients don't flow to $\epsilon$ since it was sampled from a distribution with known parameters.}\label{fig:vae_fig}
\end{figure}

Reparameterization functions denoted by T typically express the sample z in terms of its parameters $\phi$ and a random value $\epsilon$ obtained from a standard noise distribution with known parameters such that z=T($\epsilon$, $\phi$). The gradients of ELBO with respect to its parameters using the reparameterization function are given as:
\begin{equation*}
\nabla_{\theta} ELBO(\theta, \phi) = \nabla_{\theta} log P_{\theta}(x | z)
\end{equation*} 
\begin{equation*}
\nabla_{\phi} ELBO(\theta, \phi) = \nabla_{\theta} log P_{\theta}(x | z) \nabla_z \theta \nabla_{\phi} \underbrace{T(\epsilon, \phi)}_{z} - \nabla_{\phi} KL(q_{\phi}(Z)||p(Z)) 
\end{equation*}
It is sufficient to find the gradients with respect to the parameters $\theta$ and $\phi$ since the parameters of the encoder $\theta_{enc}$ and decoder $\theta_{dec}$ can be learnt by backpropagating these gradients. The figure \ref{fig:vae_fig} visualizes the gradient flow in VAE.

Reparameterization involves transforming a random sample $\epsilon$ into a random sample z from a distribution with parameters $\phi$. Largely there are two ways to construct reparameterization functions \cite{kingma2013auto}:
\begin{enumerate}
	\item \textbf{Inverse CDF}: If the distribution has a tractable inverse cumulative distribution function F$^{-1}$ then it serves as the reparameterization function with $\epsilon$ sampled from a standard uniform distribution U(0,1).
	\item \textbf{Transformations of random variables} If the distribution can be expressed as a differentiable transformation of a random variable obtained from a distribution with known parameters, then the transformation acts as the reparameterization function and $\epsilon$ is a sample from the distribution with known parameters. Eg: Location-scale families like Gaussian distribution, Laplace etc or simple composition of these families, eg: log-normal. 
\end{enumerate}
Please refer the appendix in \cite{rezende2014stochastic} for other ways to derive reparameterization functions. 

The most commonly used parametric distribution that supports reparameterization is the multivariate gaussian distribution over the dimensions of the random variable with a diagonal covariance matrix. In this case, each dimension is independent of the others with two parameters $\mu$ and $\sigma$. The reparameterization function for each dimension is:
\begin{equation}\label{reparam_gauss}
T(\epsilon, \mu, \sigma) = \mu + \epsilon \sigma
\end{equation}
Where $\epsilon$ is sampled from the standard normal distribution $\mathcal{N}$(0,1). When the approximate posterior is modelled using this distribution, the encoder outputs $\mu$ and $\sigma$ for each of the H dimensions of the latent states, this is $\phi$ that gets fed into the reparameterization function. 

Typically, VAEs are used to model images, in which case, the distribution P$_{\theta}(X|Z)$ is also modelled as a diagonal Gaussian fixing $\sigma$ to 1, so $\theta$, the output of the decoder, refers to the mean for each dimension and P$_{\theta}(X|Z)$ is directly proportional to ($\theta$ - x)$^2$. Thus, maximizing log P$_{\theta}(X|Z)$ reduces down to minimizing the mean squared error between the output of the decoder and the input image sample. 

In many cases while modelling images, the mean-squared error is called as the reconstruction loss and the KL divergence is called as the regularizer. The regularizer and reconstruction loss can be changed to improve modelling results, however, we have to be careful that any change we propose satisfies the theoretical properties of ELBO. Being over-parameterized, the network can be trained using any loss function, however, when the loss function is not an approximation of the marginal likelihood, a network behaves more like a denoising autoencoder \cite{bengio2013generalized} than a variational autoencoder. The simple objective for modelling images is:
\begin{equation*}
ELBO(\theta, \phi) = \underbrace{(\theta - x) ^ 2}_{\textit{Reconstruction loss}} \underbrace{- KL(q_{\phi}(Z)||p(Z))}_{\textit{Regularization}} 
\end{equation*}
Here $\theta$ is obtained by sampling z using \eqref{reparam_gauss} and feed it into the decoder. 
 
VAEs scale to large datasets and impose no restrictions on the complexity of P$_{\theta}(X|Z)$ allowing us to model complex data, however, only parametric distributions that have a differentiable sampling function can be used to approximate the posterior. Since the tightness of ELBO depends on the quality of this approximation, this is a huge restriction on VAEs. A large body of research is focused on mitigating these issues with VAE, this is discussed in detail in the next section.  
\newpage

\section{Research on Variational Autoencoders}
Latent variable modeling divides the problem of learning a complex distribution into a problem of learning two simple distributions, one over the latent variables and another over the data conditioned on the latent variables. Not surprisingly, improving the solution to modeling these distributions improves learning results, so a large portion of research is focused on improving these base learning models. 

A base learning model can be improved by applying any method that can learn complex probability distributions easily. Chapters \ref{improving_encoder} and \ref{improving_decoder} describe such methods and show how they applied in VAEs to improve the approximate posterior and the conditional distribution of the data. 

Chapter \ref{improving_objective} deals with the problems that arise due to approximating the marginal log likelihood and the expectation terms in ELBO. Work in this section can be used to vastly improve training in VAEs.

The differentiable sampling function allows backpropagation in  the neural networks, this is the main reason VAEs can be learnt. This technique is only applicable to a certain class of models which have continuous latent states. Chapter \ref{dealing_with_discreteness} focuses on learning VAEs in the absence of a reparameterization function, which is the case for discrete latent variable models. This chapter also highlights the importance of reparameterization. 

The following diagram summarizes the major research directions.
\begin{figure}[h]\label{vae_fig}
\center
\includegraphics[scale=0.4]{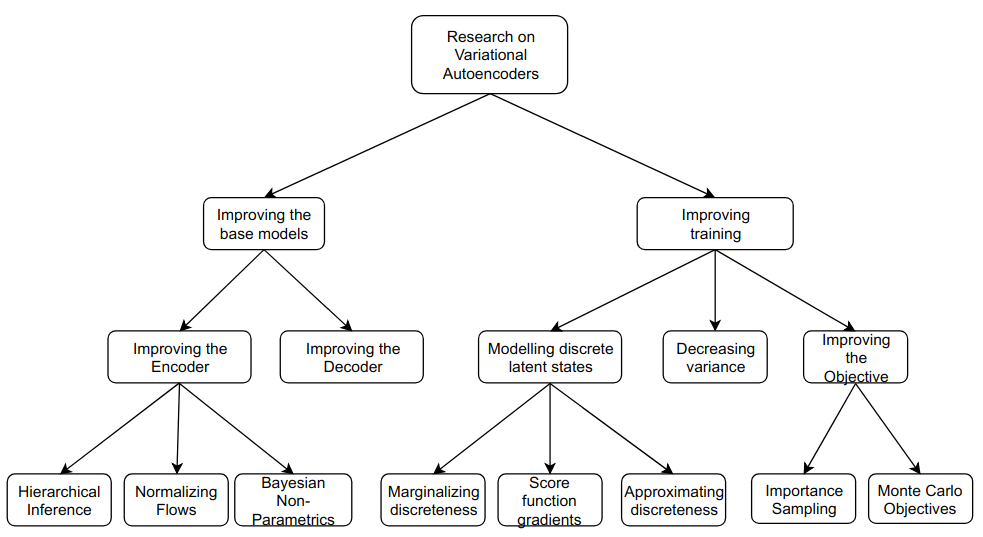}
\caption{Research on Variational Autoencoders.}
\end{figure}

\chapter{Improving the encoder} \label{improving_encoder}
The KL divergence between the approximate posterior and the true posterior dictates the tightness of the bound on the marginal likelihood as seen in \eqref{kl_posterior}. The complexity of the approximate posterior is limited by the need to keep it reparameterizable, thus if the approximate posterior is not complex enough, inference can never become equivalent to maximum likelihood estimation. In this chapter we focus on methods to increase the complexity of the approximate posterior while simultaneously keeping it reparameterizable. 

The problem of learning a complex posterior is similar to the problem of learning a complex distribution over the data, so any method that allows learning complex probability distributions can be used to improve training in latent variable models. The chapter is divided into three sections and each section discusses a method to learn complex probability distributions and shows how they are used to increase the complexity of the approximate posterior. 
\newpage

\section{Hierarchical inference}\label{hierarchical_inference}
Learning the complex distribution over the data was largely simplified by the addition of latent variables as discussed in \ref{the_problem}, we use the same idea to learn a complex approximate posterior by adding latent variables to the approximate posterior. This method was introduced by \cite{ranganath2016hierarchical} in the context of variational inference, this section discusses the method in the context of variational autoencoders.  

We can add latent variables $\lambda$ to the approximate posterior q(Z) and learn q(Z) by marginalizing over simpler distributions q(Z$| \lambda$) and q($\lambda$), this is the core idea behind hierarchical inference. The approximate posterior obtained through marginalization is given as:
\begin{equation*}
q(z) = \sum_{\lambda} \underbrace{q(z | \lambda) q(\lambda)}_{q(z, \lambda)}
\end{equation*}	

Learning the approximate posterior through marginalization becomes intractable when the number of latent states $\lambda$ is too large, this is a standard problem in latent variable models solved using VAEs. So we learn another VAE by maximizing the lower bound on the approximate posterior q(Z). This second VAE infers the posterior q($\lambda|$Z) and is seen as performing auxiliary inference. Thus, adding latent variables to latent variables creates additional inference steps, hence the name hierarchical inference.   

To learn the approximate posterior, we assume parametric distributions q$_{\phi}(Z|\lambda)$, q$_{\beta}(\lambda)$ and learn the parameters by maximizing the ELBO as:
\begin{equation*}
ELBO(\theta, \phi, \beta) = \underbrace{E_{q_{\phi, \beta}(Z)}}_{intractable}\left[log P_{\theta}(x, z)\right] - \underbrace{E_{q_{\phi, \beta}(Z)}}_{intractable}[\underbrace{log q_{\phi, \beta}(Z)}_{\textit{intractable}}]
\end{equation*}

As a first step, we reduce the number of intractable terms by replacing the expectation under q(Z) with the expectation under q(Z, $\lambda$) using the following property:
\begin{equation*}
E_{q(Z)}[f(z)] = \sum_z q(z) f(z) = \sum_z \sum_{\lambda} q(z, \lambda) f(z) = E_{q(Z, \lambda)}[f(z)]
\end{equation*}

Replacing the expectation terms gives us the objective:
\begin{equation*}
ELBO(\theta, \phi, \beta) = E_{q_{\phi, \beta}(Z, \lambda)}\left[log P_{\theta}(x, z)\right] - E_{q_{\phi, \beta}(Z, \lambda)}[\underbrace{log q_{\phi, \beta}(Z)}_{\textit{intractable}}]
\end{equation*}

$log q_{\phi, \beta}(Z)$ is made tractable by approximating the intractable posterior q($\lambda|Z)$ with a parametric distribution r$_{\beta_{aux}}$($\lambda$) as follows:
\begin{equation*}
log q_{\phi, \beta}(z) = log q_{\phi, \beta}(z, \lambda) - \underbrace{log q(\lambda | z)}_{\textit{Source of intractability}} = log q_{\phi, \beta}(z, \lambda) - log r_{\beta_{aux}}(\lambda)
\end{equation*}

This gives us the objective:
\begin{equation*}
ELBO(\theta, \phi, \beta, \beta_{aux}) = E_{q_{\phi, \beta}(Z, \lambda)}\left[log P_{\theta}(x, z) - log q_{\phi, \beta}(z, \lambda) + log r_{\beta_{aux}}(\lambda)\right]
\end{equation*}

Learning the parameters $\theta, \phi, \beta, \beta_{aux}$ requires maximizing the ELBO. Applying the VAE philosophy, the expectation term is replaced with the Monte Carlo approximation and the optimizations are replaced with neural networks. The objective obtained after replacing the expectation terms is:
\begin{equation*}
ELBO(\theta, \phi, \beta, \beta_{aux}) = \frac{1}{S} \sum_{s=1}^S \left[log P_{\theta}(x|z_s) + log P(z_s) - log q_{\phi, \beta}(z_s, \lambda_s) + log r_{\beta_{aux}}(\lambda_s)\right]
\end{equation*}
Here P(Z) is the prior chosen by the practitioner, S is the total number of samples, $\lambda_s$ is a obtained by sampling from q$_{\beta}(\lambda)$. z$_s$ is the random value obtained by sampling from q$_{\phi}(Z|\lambda_s)$. $\epsilon_1$ and $\epsilon_2$ are random variables obtained by sampling from a noise distribution. Typically, S=1 is used.

\begin{figure}[h]\label{fig:hierarchical_inference}
\center
\includegraphics[scale=0.4]{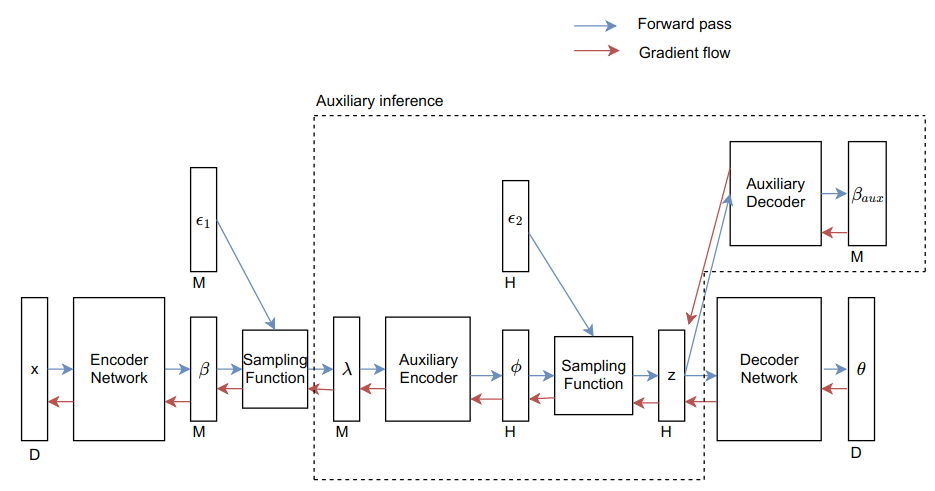}
\caption{Architecture of Hierarchical inference. H, M and D represent the dimensionality of the vectors}
\end{figure}

The encoder network takes the input x and outputs the parameter $\beta$, a value of $\lambda$ is sampled from q$_{\beta}(\lambda)$ and fed into another network that takes $\lambda$ as inputs and outputs the parameters $\phi$. The latent state z is sampled from q$_{\phi}$(Z$|\lambda$) and fed as input to two networks, one that outputs the parameters $\theta$ modelling the data and another that outputs $\beta_{aux}$ the parameters of the auxiliary posterior. The architecture of this network along with the gradient flow can be seen in \ref{fig:hierarchical_inference}.

Hierarchical inference increases the complexity of the approximate posterior, however, the objective it maximizes is a looser bound on the marginal log likelihood than the traditional ELBO. While the method strives to increase the complexity of the posterior to tighten the bound, it works by defining a looser bound on the marginal log likelihood, so it is hard to know if the tightness brought about by the complex posterior would compensate for the looseness introduced to learn it. Evaluating the tightness of the bound is intractable since the marginal likelihood itself is intractable, hence there are no theoretical guarantees that support hierarchical inference methods. Empirical results from \cite{ranganath2016hierarchical} seem to indicate that they do perform better. 

In the next section we look into normalizing flows which allow us to increase the complexity of the posterior without loosening the bound on the marginal log likelihood.  
\newpage

\section{Normalizing flows}\label{normalizing_flows}
A normalizing flow is a general technique to learn the probability distribution of data through maximum likelihood estimation. It models the distribution of the data by mapping the data to a subspace and learning the distribution of the data in the subspace. In this section, we describe normalizing flows and show how they can be used to improve the complexity of the approximate posterior.

Learning the distribution of the data through maximum likelihood estimation(MLE) requires knowing the log probability of the data. So any method that performs MLE by breaking down learning into smaller sub-problems should be able compute the log probability of the data. Latent variable models compute it using marginalization and normalizing flows apply the change of variables technique. As the size and complexity of the data increases, marginalization becomes intractable and requires approximations, the change of variables technique on the other hand stays tractable allowing normalizing flows to perform MLE on large complex datasets. 

The change of variables technique is a standard method to obtain the probability of a random variable before transformation in terms of the probability of the transformed variable. Assume x represents the data sample, f the transformation function and p$_s$ the distribution of the data in the subspace, using the change of variables technique we can express logp(x) in terms of p$_s(f(x))$ and f as:
\begin{equation}\label{eq:change_of_variable}
log p(x) = log p_s(f(x)) + log | det \frac{\delta f(x)}{\delta x} |
\end{equation}
Here $|det \frac{\delta f(x)}{\delta x}|$ refers to the absolute value of the determinant of the Jacobian of the transformation function. 

For learning, we assume a parametric transformation function f$_{\alpha}$ and a parametric distribution in the subspace, p$_{\theta}$. We learn these parameters $\alpha$ and $\theta$ by maximizing the log probability over the entire dataset. 

Research in normalizing flows is focused on defining parametric transformation functions $f_{\alpha}(x)$ for which $| det \frac{\delta f_{\alpha}(x)}{\delta x} |$ can be easily computed. \cite{dinh2014nice} and \cite{dinh2016density} use properties of the determinant to define these transformations and demonstrate good results on image modeling. 

In MLE, we are interested in simplifying the distribution of the data to make learning easy. On the contrary, in latent variable models we are interested in increasing the complexity of the distributions to improve learning capacity. So instead of mapping random variables from a complex subspace to a simple subspace as done in MLE, we map variables from a simple subspace to a complex subspace. 

Normalizing flows model complex approximate posteriors by mapping random variables from a simple approximate posterior to a complex subspace. Assume z$_0$ represents a random value sampled from the approximate posterior, q$_{\phi}(Z_0)$ and f$_{\alpha_1}$ the parametric transformation function, we can obtain the probability in the complex subspace, q$_{nf}(f_{\alpha_1}(z_0))$ using \eqref{eq:change_of_variable} as:
\begin{equation*}
log q_{nf}(f_{\alpha_1}(z_0)) = log q_{\phi}(z_0) - log | det \frac{\delta f_{\alpha_1}(z_0)}{\delta z_0} |
\end{equation*}

Let z$_1$ denote the transformed random variable such that z$_1=f_{\alpha_1}(z_0)$. A transformation can be specified on z$_1$ to obtain a transformed variable z$_2$, in this fashion, we can obtain a random variable z$_k$ by applying K transformation functions as:
\begin{equation*}
z_k = f_{\alpha_k}(f_{\alpha_{k-1}}(.....(f_{\alpha_1}(z_0)...)
\end{equation*}

The probability of z$_k$ can be specified using the following equation:
\begin{equation*}
log q_{nf}(z_k) = log q_{\phi}(z_0) - \sum_{k=1}^K log | det \frac{\delta f_{\alpha_k}(z_{k - 1})}{\delta z_{k - 1}} |
\end{equation*}

The random value z$_k$ obtained after applying the series of transformations is fed into the decoder to model the conditional distribution of the data. The distribution $q_{nf}(Z_k)$ from which z$_k$ is obtained acts as the new approximation to the posterior. ELBO in terms of this new distribution is given as:
\begin{equation*}
ELBO(\theta, \phi, \alpha) = E_{q_{nf}(Z_k)}\left[log P_{\theta}(x| z_k) + log P(z_k) - log q_{\phi}(z_0) + \sum_{k=1}^K log | det \frac{\delta f_{\alpha_k}(z_{k - 1})}{\delta z_{k - 1}} |\right]
\end{equation*}
As always, the prior P(Z$_k$) is chosen by the practitioner. 

\begin{figure}[h]\label{fig:normalizing_flows}
\center
\includegraphics[scale=0.4]{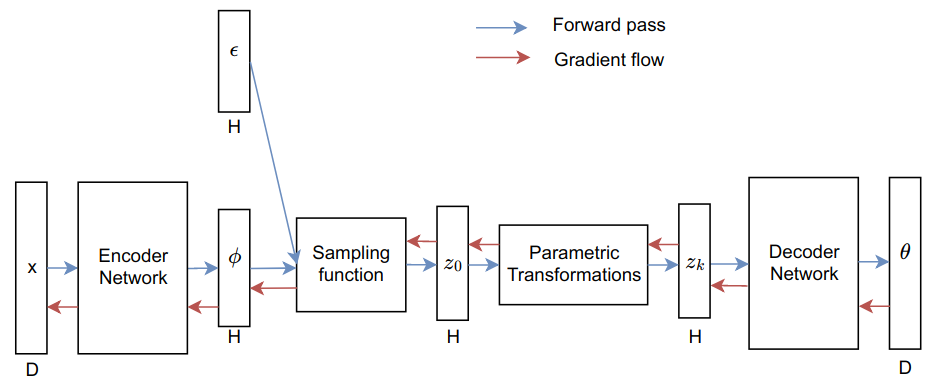}
\caption{Architecture of Parametric transformations(Normalizing flows)}
\end{figure}

Since the distribution $q_{nf}(Z_k)$ allows sampling by sampling from q$_{\phi}(z_0)$ which inturn has a differentiable sampling function, the expectation term in ELBO can be replaced with Monte-Carlo approximation to obtain the objective:
\begin{equation*}
ELBO(\theta, \phi, \alpha) = \frac{1}{S} \sum_{s=1}^S \left[log P_{\theta}(x| z_k) + log P(z_k) - log q_{\phi}(z_s) + \sum_{k=1}^K log | det \frac{\delta f_{\alpha_k}(z_{k - 1})}{\delta z_{k - 1}} |\right]
\end{equation*}
Here S is the number of samples, z$_s$ is sampled from the simple approximate posterior q$_{\phi}(Z_0)$ and z$_k$ is obtained by passing z$_s$ through the series of transformation functions. The architecture of VAE using normalizing flows is shown in \ref{fig:normalizing_flows}.

There are a large number of transformation functions that can be used in normalizing flows, however, the most popular ones are  inverse autorgressive flow \cite{kingma2016improved} and Non Linear Independent Component Estimation \cite{dinh2014nice}. Several other transformation functions can be found in \cite{tomczak2016improving}, \cite{rezende2015variational}

The number of transformations used in a normalizing flow is called as the length of the flow, for certain transformation functions, we can analyze the effect of applying transformations on the initial distribution q$_{\phi}(Z_0)$ as a function of the length of the flow. Infinitesimal flows can be constructed that converge to the true posterior when the flow length becomes infinity. Construction and analysis of infinitesimal flows require a basic understanding of fluid dynamics, more details can be found in \cite{rezende2015variational}. 

In this section we discussed normalizing flows as a way to increase the complexity of the approximate posterior without further lower bounding ELBO as done in hierarchical inference. Normalizing flows can also be coupled with hierarchical inference by using it to improve the complexity of any latent hierarchy. It can also be used to improve the complexity of the conditional distribution of the data modelled by the decoder. 

The complexity of the transformation functions in normalizing flows are limited by the need to keep $| det \frac{\delta f_{\alpha}(x)}{\delta x} |$ tractable. Volume preserving flows\cite{dinh2014nice} are a special class of normalizing flows that sidestep this issue by constructing transformation functions with unit determinant, however they are not as effective. In the next section, we discuss variational gaussian processes, they combine ideas from normalizing flows and hierarchical inference and learn arbitrarily complex posteriors without the computational bottleneck.
\newpage

\section{Bayesian Non-parametrics}
Bayesian inference learns the distribution of the data by assuming a parametric distribution over the data and treating the parameters of the distribution as random variables. It assumes a distribution over the parameters called the prior that it learns based on the data. Typically, the domain of the parameters are restricted by distribution of the prior. A special case is the Bayesian Non-parametric model where the prior places no restrictions on the parameters allowing the model to learn arbitrarily complex distributions. In this section we describe Gaussian Process regression and show how they can be used to construct a Bayesian non-parametric model. We later discuss how they can be  used in VAEs to learn complex approximate posteriors. This method was proposed by \cite{tran2015variational} as the Variational Gaussian Process.

\subsection{Gaussian Process Regression}
Given a dataset of m source-target pair vectors (s, t) where s denotes a C dimensional source vector and t the corresponding H dimensional target vector, a regression problem involves learning a transformation function f such that t=f(s). A Gaussian Process(GP) is a distribution over such transformation functions that map a given input vector to an output vector. It models the functions by modeling the output of the function on a particular input, so it is represented as a multivariate gaussian where the mean and covariance matrices are functions of the input, denoted as $\mathcal{GP}$(f; mean(s), cov(s)).  

For a given input, a gaussian process models the output by interpolating the target corresponding to the closest source vector present in the data. Here closeness between a source vector s and an input vector s' is measured using a kernel function k(s, s') specified by the practitioner. A simple choice is the automatic relevance detection kernel given as:
\begin{equation*}
k(s, s') = \sigma_{ard}^2 exp(-\frac{1}{2}\sum_{i=1}^C w_i (s_i - s_i') ^ 2)
\end{equation*}
Here $\sigma_{ard}$ and w$_i$ are parameters of the kernel function. 

Regression using a Gaussian process is performed by assuming a prior over the transformation functions and learning the posterior distribution of the functions conditioned on the data, the set of m source-target pairs. Assuming each dimension of the output is modeled using an independent gaussian process, the prior over the functions is given as:
\begin{equation}\label{gp_prior}
p(f) = \prod_{i=1}^H \mathcal{GP}(f_i;0, K_{ss})
\end{equation}
Here K$_{ss}$ denotes the covariance function evaluated over all the source pairs using the kernel function k(s, s') and f$_i$ represents $i^{th}$ dimension of the target which is a mapping from the entire source vector to a single vector i.e R$^C$ $\rightarrow$ R.

The posterior distribution of the functions evaluated at the input $\epsilon$ a C dimensional vector, given the data D=(s,t) the set of source target pairs is:
\begin{equation}\label{gp_conditional}
p(f | D, \epsilon) = \prod_{i=1}^H \mathcal{GP}(f_i;K_{\epsilon s}K_{ss}^{-1}t_i, K_{\epsilon \epsilon}K_{ss}^{-1}K_{\epsilon s}^T)
\end{equation}
Here t$_i$ is the i$^th$ dimension of the target over the entire dataset. The derivation of this result can be found in \cite{38136}. It is easy to see that the complexity of this method scales with the number of data points, which it is a non-parametric of learning. 

\subsection{A Bayesian Non-parametric Model}
Section \ref{normalizing_flows} demonstrated how random variables in a simple subspace can be mapped onto a complex subspace, similarly, the parameters of the distribution can be obtained by mapping random variables from a simple distribution. The set of parametric transformation functions that can be specified to learn this mapping however is restricted by tractability constraints, this also limits the domain of the parameters. Gaussian processes lifts this restriction by learning a distribution over the transformations instead, hence they can be used as a prior to obtain bayesian non-parametric models. This subsection discusses the construction of such a model and shows how they can be used to learn distributions over data.

The source and target pairs required by the Gaussian process to learn the distribution over transformations are treated as parameters, this way, the pairs adapt to the problem at hand allowing the GP to learn distributions over arbitrarily complex transformations. The parameters of the kernel function and the source-target pairs are learnt through optimization.

Let x represent the observed data sample. To learn the distribution over X, we assume a simple parameteric distribution P$_{\theta}$(X) such that $\theta$ is a random variable obtained by transforming $\epsilon$ from the standard normal $\mathcal{N}$(0, I). We assume a gaussian process to describe the distribution over these transformations with parameters $\lambda$ = $\{\sigma_{ard}, w_1, .. w_H, (s_1, t_1)...(s_m, t_m)\}$ which is the set of kernel parameters and the source target pairs data. We can express the log probability of the observed data as follows:
\begin{equation*}
log P(x) = log \int_{\epsilon} \int_{f} P(x; f(\epsilon)) P(f, \epsilon) df d\epsilon
\end{equation*}
Here $\theta$ = f($\epsilon$) and $P(f, \epsilon)$ denotes the joint distribution over the functions and the random noise. Here the functions and $\epsilon$ act as latent variables for the probability distribution. It is clear that the marginalization is intractable so we use methods from latent variable modeling to construct a lower bound that allows optimization and learn the parameters $\lambda$ by optimizing the lower bound. 

By assuming a distribution q(f, $\epsilon$) over the latent variables, we obtain the objective which is a lower bound on logP(x) as:
\begin{equation*}
objective(\lambda) = E_{q(f, \epsilon; \lambda)}\left[logP(x; f(\epsilon)) + log P(f, \epsilon) - log q(f, \epsilon; \lambda) \right]
\end{equation*}
Here $P(f, \epsilon)$ is the prior decided by the practitioner. Typically, we assume independent distribution over the functions and the noise for the prior. The GP prior \eqref{gp_prior} over functions is chosen for P(f) and a standard normal distribution for P($\epsilon$) and q($\epsilon$). We assume q(f$|$ $\epsilon; \lambda$) as the conditional distribution over the functions given the input $\epsilon$ as seen in \eqref{gp_conditional} and data in $\lambda$. This gives us the objective:
\begin{equation*}
objective(\lambda) = E_{q(f, \epsilon; \lambda)}\left[logP(x; f(\epsilon)) + log P(f; \lambda) - log q(f|\epsilon; \lambda) \right]
\end{equation*}
As long the choice of the parametric distribution P$_{\theta}(X)$ allows differentiation with respect to its parameters, the above function can be optimized by replacing the expectation term with Monte Carlo approximations since $q(f, \epsilon; \lambda)$ allows sampling by first sampling $\epsilon$ from a standard normal and then using that to sample f using reparameterization trick on the gaussian process. The differentiable sampling for the gaussian process is given as:
\begin{equation*}
f_i(\epsilon_1; \lambda) = L\epsilon_1 + K_{\epsilon_0 s}K_{ss}^{-1}t_i
\end{equation*}
Here f$_i$ denotes the i$^{th}$ dimension of the parameters modeled by the GP and L denotes the cholesky decomposition of the covariance $K_{\epsilon_0 \epsilon_0}K_{ss}^{-1}K_{\epsilon_0 s}^T$ evaluated at the input noise $\epsilon_0$. Here $\epsilon_1$ is also sampled from a standard normal, many times, it is common to use the same noise sample $\epsilon_0$ which is used to compute the mean and covariance to also sample from the multivariate GP, in which case $\epsilon_1$ = $\epsilon_0$.

The objective in terms of the Monte-Carlo approximation is given as:
\begin{equation*}
objective(\lambda) = \frac{1}{S} \sum_{s=1}^S \left[logP(x; f_s(\epsilon_s)) + log P(f_s; \lambda) - log q(f_s|\epsilon_s; \lambda) \right]
\end{equation*}
Here f$_s$ and $\epsilon_s$ denote the s$^{th}$ sample, not to be confused with the s$^{th}$ dimension. 

Since the objective is differentiable with respect to $\lambda$ the parameters of the kernel function and the source target pairs are learnt based on the data. Since there are no restrictions on the source-target pairs, the GP learns arbitrarily complex transformations, thereby learning complex distributions over the input x.

\begin{figure}[h]\label{fig:bayesian_non_param}
\center
\includegraphics[scale=0.3]{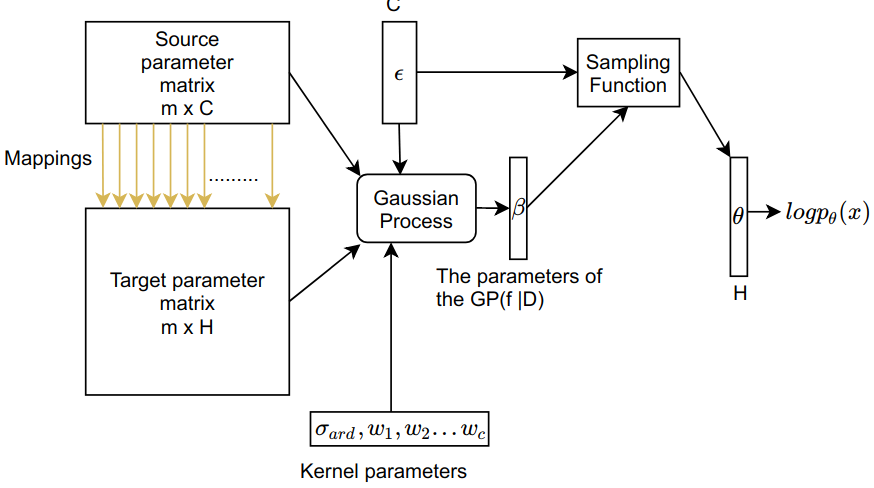}
\caption{Architecture of the Bayesian non-parametric model constructed using GP.}
\end{figure}

\subsection{Variational Gaussian Process}
The previous subsection showed how a Bayesian non-parametric model can be trained to learn the complex distribution over the data. Here, we use the model to learn a complex approximate posterior by using principles from hierarchical inference. A more detailed description of this method can be found in \cite{tran2015variational}.   

Recall that the Bayesian non-parametric model defined in the previous section is a latent variable model where the GP along with the standard normal defines the distribution over the latent variables and $\lambda$ is the set of parameters of the latent distribution. The core idea of VGP is to learn the approximate posterior q(z) as a latent variable model defined by the Bayesian non-parametric model. In order to learn such a model it applies the auxiliary inference technique discussed in section \ref{hierarchical_inference}. 

The distribution of the approximate posterior using a Bayesian non-parametric model is given as:
\begin{equation*}
q(z; \lambda) = \int_{\epsilon} \int_{f} q(z; f(\epsilon)) q(f| \epsilon; \lambda)\mathcal{N}(\epsilon| 0, I) df d\epsilon
\end{equation*}

As long as the parametric distribution q(Z) allows reparameterization and is differentiable with respect to its parameters, the hierarchical inference method can be applied directly. 

\begin{figure}[h]\label{fig:vgp}
\center
\includegraphics[scale=0.4]{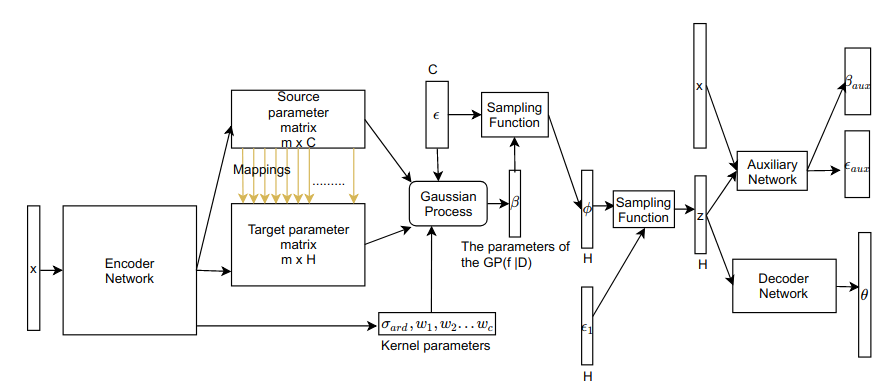}
\caption{Architecture of the Variational Gaussian Process.}
\end{figure}

In the context of hierarchical inference, $\lambda$ is the set of parameters of the second level latent distribution, so such parameters are modeled using the encoder which takes the data sample x as input and outputs the source-target pair data and the parameters of the kernel function which together makes up $\lambda$. Based on $\lambda$ the GP learns the mean and covariance over the transformations as a function of $\epsilon$. $\epsilon$ is sampled from the standard normal and fed into the VGP to sample a function f from the GP. f in combination with $\epsilon$ is used to obtain f($\epsilon$) the parameters of the approximate posterior. Since f and $\epsilon$ are latent variables of the latent distribution, auxiliary inference is setup and an auxiliary distribution r(f, $\epsilon$ $|$ z, x) is learnt using a separate neural network. The architecture of entire VGP is as shown in \ref{fig:vgp}.

The VGP is able to model complex approximate posteriors without the computational burden of being able to compute the determinant. The method however uses auxiliary inference to learn as result of using hierarchical inference. Similar to infinitesimal flows, they have theoretical results that demonstrate that in the presence of infinite-source target pairs data, it can model the true posterior over the latent variables. 

\chapter{Improving the decoder} \label{improving_decoder}
The decoder learns the distribution of the data conditioned on the latent information, since it is a sub-problem in latent variable modeling, we can expect improvement in results by making it easier for the decoder to learn complex conditional distributions. However, the distribution cannot get arbitrarily complex since if the decoder can model the data without the latent information, it will not pass any learning signal to the encoder. The distribution modeled by the encoder controls how close training gets to maximum likelihood estimation, for this  reason it is crucial to ensure that the decoder does not get too complex. 

Similar to methods that improved the distribution of the encoder, any method that can learn complex probability distributions can be used to improve the decoder, so a method that improves the encoder can also improve the decoder. In this chapter, we discuss how autoregressive models can be used to improve the complexity of the decoder. As always, we start by discussing how autoregressive models can be used to learn complex distributions and later show how they improve the decoder.  

When modeling multidimensional data samples, it is common to model each dimension of the sample independent of the other dimensions. However, such methods ignore information contained in the neighboring dimensions. In many data sets like images and audio, there is a high degree of dependence between neighboring dimensions, for eg: we can expect a pixel to have the same color in natural images if all its neighbors have the same color. Autoregressive models exploit this dependency by constructing the distribution of a dimension conditioned on its neighbors. 

Consider a D dimensional data sample x, assume that there is a particular ordering among these dimensions such that every i$^{th}$ dimension can be said to be dependent on its previous k dimensions $\{i-k,...i-1\}$. We can express the probability of the sample x using an autoregressive model as:
\begin{equation*}
p_{\theta}(x) = p_{\theta_1}(x_1)p_{\theta_2}(x_2 | x_1, x_0)....p_{\theta_i}(x_i | x_{i-1} ..,x_{i-k})...p_{\theta_D}(x_D | x_{D-1} ...,x_{D-k})
\end{equation*}
Here $\theta_1$..$\theta_D$ represents the parameters for each dimension stored collectively in a parameter vector $\theta$ = $[\theta_1..\theta_D]$. 

An autoregressive model can be constructed using a recurrent neural network where each timestep of the network is used to model a particular dimension. At each timestep, the network takes k neighboring dimension as input and outputs the distribution over the particular dimension. In such a construction the parameters of each dimension of the data is amortized by using a single large parametric recurrent network. Since the log probability of the network can be computed using the output distribution at each timestep, the parameters of the network can be learnt by maximizing the log probability of the data, i.e maximum likelihood estimation. The first k dimensions of the input are modeled by feeding zeros as the input to its neighbors. 

Each dimension of the input has to be modeled sequentially in autoregressive models, this makes modeling and sampling extremely slow for high dimensional data. It is also hard to decide the ordering of the dimensions that yields the best representation of the input, \cite{gregor2015draw} one method to learn the ordering by learning a selective attention mechanism that shows good results on image modeling. 

Integrating autoregressive models into the decoder is straightforward, the recurrent neural network modeling the conditional distribution is designed to take an additional vector, the latent vector, as input. Each dimension of the data is thus modeled by feeding the neighboring dimensions along with the latent state sampled from the encoder. Traditionally, VAEs assume the data dimensions to be independent of each other. \cite{gulrajani2016pixelvae}, \cite{gregor2015draw} and \cite{chen2016variational} demonstrate state of the art results on image modeling when autoregressive models are used to learn the conditional distribution of the data. 

The general idea to control the complexity of autoregressive models in VAEs is to design the autoregressive model over the data in such a way so that it cannot model the distribution we want the latent states to model but can accurately model the distribution that the latent states cannot model. Such networks have to be carefully constructed based on the data we are trying to model. Variational lossy autoencoders \cite{chen2016variational} give a particular example of constructing autoregressive convolutional recurrent networks that can only model the local texture in the image forcing the latent states to capture global structure. 

To empirically test if a decoder ignores latent information, latent information from different data samples can be fed into the decoder and the change in the output for different samples can be monitored. Such tests only provide a rough indication, a valuable direction of research in this space would be to construct theoretical techniques to measure the utilization of the latent states using ideas from information theory. 

Another way to increase the complexity of the decoder is by incorporating ideas from Generative Adversarial Networks \cite{goodfellow2014generative}. Here, the decoder in VAE is can be made to output the data instead of the parameters of the distribution, the probability of the data output by the decoder is learnt using a discriminator network that is trained to distinguish between real images and the images from the decoder. More details on this construction can be found \cite{larsen2015autoencoding}. We do not discuss this since learning adversaries do not provide fine control over the complexity of the distribution which is required to ensure that the latent states are not ignored.  

\chapter{Modelling Discrete Latent states} \label{dealing_with_discreteness}
Discrete latent states can be used to capture many useful representations of data, for example the number of objects in an image, the number of chords in a song etc and they can also be useful when learning downstream tasks as in Reinforcement learning. VAEs however cannot be used to learn discrete latent states over the data due to lack of a differentiable sampling function. This section discusses approaches to train VAE when the latent states don't have differentiable sampling functions. 
\\

The operation of sampling in discrete latent states involves mapping the continuous parameters of the discrete distribution to discrete samples, any function that maps from continuous space to a discrete space cannot be inverted, for this reason distributions over discrete latent states do not have a differentiable sampling function. 

The following are dominant approaches to working with discrete latent states:
\begin{enumerate}
	\item Marginalizing discreteness
	\item Score function
	\item Approximating discreteness
	\item Parameterizing discreteness
\end{enumerate}
\newpage

\section{Marginalizing discreteness}
When there are a small number of discrete latent states, the best approach is to assume a set of continuous latent states along with the discrete states and marginalize out the discrete states. The distribution obtained after marginalizing is continuous and as long as we choose continuous distributions that support reparameterization, this trick can be used. 
\\

Consider the latent state vector Z to be comprised of two parts, Z$_1$, the continuous part and Z$_2$, the discrete part. Assume that the two parts are independent of each other such that any distribution over the latent states q$_{\phi}$(Z$_1$, Z$_2$) can be expressed as the product over the individual distributions as:
\begin{equation*}
q_{\phi}(Z_1, Z_2) = q_{\phi}(Z_1)q_{\phi}(Z_2)
\end{equation*}

In this scenario, we can express the expected value of any function under the distribution over the latent states as:
\begin{equation*}
E_{q_{\phi}(Z_1, Z_2)}\left[f(x, z_1, z_2)\right] = E_{q_{\phi}(Z_1)}\left[E_{q_{\phi}(Z_2)}\left[f(x, z_1, z_2)\right]\right]
\end{equation*}

When the number of latent states of Z$_2$ is small, the expectation $E_{q_{\phi}(Z_2)}\left[f(x, z_1, z_2)\right]$ can solved in closed form and Monte-Carlo approximations of this closed form gradients can be used by obtaining continous samples from q(Z$_1$). 

Thus in this method, the only change is in the objective function where the discrete states are marginalized out. The architecture of the network is such that the encoder outputs a distribution over the latent states Z and the decoder learns the conditional distribution of the data based on both the discrete and the continuous latent states in Z. 

Another variant of marginalizing discreteness is studied in \cite{rolfe2016discrete}, it is very similar to the method discussed, however, it performs the marginalization by using a conditional cumulative distribution function instead of the complete distribution function.
\newpage

\section{Score function gradients}
The reparameterization function exposes a differentiable function between the output of the encoder and the samples z. By treating z as an input instead of a result from a sampling function, score function gradients allow optimization of discrete latent states in VAE.
\\

The method uses the property that the derivative of a function can be expressed as the product of the function times its log derivative as follows:
\begin{equation*}
\nabla q(Z) = q(Z) \nabla log q(Z)
\end{equation*}

The above result allows us to express the gradient of the ELBO with respect to the parameters of the encoder $\theta_1$ as:
\begin{equation*}
\nabla_{\theta_1} ELBO = \sum_{z} \left[log P(x, z) - log q(z; \theta_1) \right]\nabla_{\theta_1} q(z)
\end{equation*}
\begin{equation*}
= E_{q(z)}\left[(log P(x, z) - log q(z; \theta_1)) \nabla_{\theta_1} log q(z; \theta_1)\right]
\end{equation*}
A detailed derivation can be found in the appendix.

The gradient with respect to the parameters $\theta_2$ of the decoder can be obtained as:
\begin{equation*}
\nabla_{\theta_2} ELBO = E_{q(z)} \left[ \nabla_{\theta_2} log P(x, z; \theta_2) \right]
\end{equation*}

Monte Carlo approximations of these expectations using samples from q(z) can be used to learn the parameters of the encoder and the decoder that maximize the ELBO. 

Since the method ignores the relationship between the output of the encoder and the samples z, it cannot account for the variance between each of the samples z obtained from q(z), this introduces  variance among the gradients estimated by each of the samples. When this variance is high, there is considerable difference between the gradient estimates of each sample leading to poor optimization. The main reason VAEs are able to learn complex distributions with continuous latent states is because of the differentiable sampling function which allows us to obtain low variance gradient estimates of ELBO.
\\

High variance is the price paid by score function estimators to allow optimization of discrete latent states. To decrease the variance, control variates are commonly used. A control variate is a function that is correlated with the objective function such that when added to the objective it does not alter the magnitude of the objective but decreases its variance. Appendix, section \ref{control variates} discusses control variates in greater depth.

Hierarchical inference and mean field distributions can be combined to decrease the variance in score function gradient estimators. Details of this method can be found in \cite{ranganath2016hierarchical}.
\newpage

\section{Continuous approximations of discreteness}
A large body of research in this field is based on the idea of approximating bottlenecks. Applying the same idea, we approximate the discrete latent variables using continuous latent variables that have a differentiable sampling function.
\\

In these methods, the discrete latent variable is represented as a one hot encoded vector with a parameter vector $\alpha$ such that the k$^{th}$ dimension $\alpha_k$ represents the unnormalized probability of the k$^{th}$ discrete state. In the VAE framework, the encoder outputs the parameters $\alpha$ to model discrete latent states and a discrete state z$_d$ can be sampled using the Gumbel-max trick, given as:
\begin{equation*}
z_d = argmax (log \alpha_i - log(-log U_i))
\end{equation*}
Here the argmax is taken over all discrete states, U$_i$ represents a random sample from the uniform distribution U(0, 1) and -log(-log U) is commonly known as the Gumbel distribution. 
\\

Using continuous approximations allows us to estimate the gradient of the ELBO required for training. There are broadly two ways in which continuous approximations can be used and they differ in the nature of their gradients estimates, they are:
\begin{enumerate}
	\item Biased estimates of gradient
	\item Unbiased estimates of gradient
\end{enumerate}

\subsection{Biased estimates of gradients}
The core idea of these methods is to replace discrete random variables and their distributions with the corresponding continuous approximations. This makes learning biased towards the continuous approximations, however the degree of the bias can be usually controlled based on the approximating function. Although the learning is biased, the use of reparameterization function ensures low variance. 
\\

For a function to serve as a continuous approximation, it should take the parameters $\alpha$ and some random noise $\epsilon$ as input and output a continuous random variable z$_c$ that closely approximates the discrete random variable z$_d$. If the continuous approximation is represented by f$_c$ with parameters $\lambda$, the function is given as:
\begin{equation*}
z_c = f_c(\alpha, \epsilon; \lambda)
\end{equation*}

The method works through a process called relaxing, which involves replacing all the discrete random variables z$_d$ by continuous random variable z$_c$. In the context of VAE, the encoder outputs the parameters $\alpha$, this is used by the function f$_c$ to obtain z$_c$ which is fed into the decoder. If z$_c$ is a good approximation of z$_d$, the effect of feeding z$_c$ into the decoder should be similar to feeding z$_d$, this ensures that parameters learnt in the continuous space perform well in the discrete space. 
\\

From an implementation perspective, relaxing the variables is sufficient to allow learning, from a theoretical perspective however, ELBO will not be defined since continuous variables z$_c$ from a distribution q(Z$_c|$X) is fed into the decoder. So the expectation terms in ELBO and log probability of the approximate posterior should be replaced with the density of the continuous approximation. A common mistake while building VAEs is to not recognize this problem and train the networks on the same objective, recall that such a network still trains but behaves more like a denoising autoencoder than a variational autoencoder. 

\begin{figure}[h]\label{fig:approx_discrete}
\center
\includegraphics[scale=0.3]{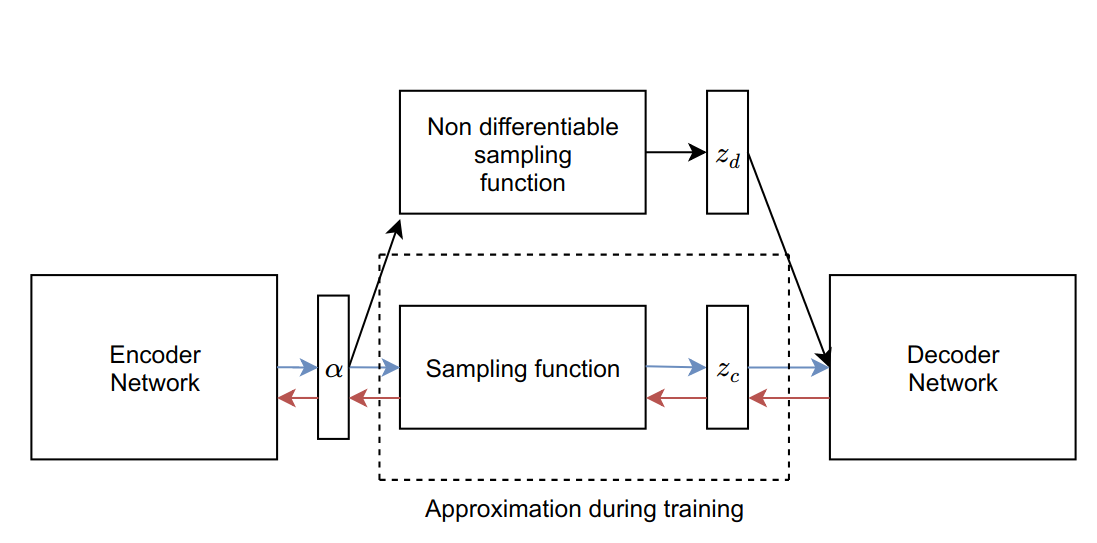}
\caption{Biased estimation with discrete latent states.}
\end{figure}

For theoretical support, the probability density over the continuous random variable q(Z$_c|$X) will have to be determined and used, this process of modifying the ELBO by replacing the discrete probability density with the continuous probability density is called relaxing the objective. Therefore, by relaxing the variables and the objective, VAEs can be trained with discrete latent states. The architecture of the VAE with relaxing is shown in \ref{fig:approx_discrete}.

To summarize, f$_c$ should output a random variable whose probability density can be easily obtained and is a close approximation of the discrete random variable z$_d$. One such function is the Gumbel softmax function given as:
\begin{equation*}
f_{ck}(\alpha_k) = \frac{exp(\frac{log \alpha_k + G_k}{\lambda})}{\sum_{i=1}^n exp(\frac{log \alpha_i + G_i}{\lambda})}
\end{equation*}
f$_{ck}$ represents that the function models the k$^{th}$ dimension of the output, G$_i$ is a sample from the Gumbel distribution -log(-log U$_i$) obtained by sampling a random variable from U(0,1) and evaluating the Gumbel function, $\lambda$ is a parameter that controls how close the distribution is to the discrete distribution, as $\lambda$ $\rightarrow$ 0, it approaches the discrete distribution. The relation between z$_c$ and z$_d$ can be expressed as:
\begin{equation*}
z_d = H(z_c) = one\_hot(argmax(z_{c}))
\end{equation*}

The probability density of the gumbel softmax can be derived using the change of variables technique on the gumbel distribution and is given as:
\begin{equation*}
q(Z_c | X) = (n-1)! \lambda^{n - 1} \prod_{k=1}^n \left(\frac{\alpha_k z_{ck}^{-\lambda - 1}}{\sum_{i=1}^n \alpha_i z_{ci}^{-\lambda}}\right)
\end{equation*}
Here n denotes the number of discrete latent states.
\\

The hyperparameter $\lambda$ of the Gumbel softmax controls how close the distribution is to the discrete distribution, thereby controlling the bias of learning. A high value of $\lambda$ implies that the VAE will be more biased towards the continous approximation of the latent state than the discrete latent state, however a low value of $\lambda$ will result in high variance in the gradient estimates of the samples. Typically, a value of $\lambda$ close to n - 1 is chosen at the start of training since it has good theoretical properties \cite{maddison2016concrete}. The value of $\lambda$ is periodically decreased as training progresses. 

\subsection{Unbiased estimates of gradients}
Relaxing the objective allows training however the training is biased towards the continuous latent states. In this section, we discuss REBAR \cite{tucker2017rebar} a method that allows us to construct control variates using the biased gradient estimator and the score function gradient estimator. This control variate is added to the objective to obtain unbiased estimates of the gradients.
\\

To avoid excessive clutter, assume L(x, z) = $\left(log \frac{P(x, z)}{q(z; \theta_1)}\right)$, a suitable control variate for score function gradient can be written as:
\begin{equation} \label{eq:control_variate}
\nabla_{\theta_1} ELBO = E_{q(z)}\left[L(x, z)\nabla_{\theta_1} log q(z; \theta_1)  - c\right] + E_{q(z)}[c]
\end{equation}
Here c represents the control variate which can either depend on the random variable z or be constant, and has known expectation $E_{q(z)}[c]$. When $E_{q(z)}[c]$ is not known, a low variance gradient estimate such as the one obtained through reparameterization can be used. 
\\

The control variate is designed by recognizing that the score function gradient of the continuous latent state is equal in magnitude to its reparameterization gradient so adding the score function gradient and subtracting the reparametrization gradient leaves the objective with respect to the discrete latent states unchanged. The score function gradient of the continuous latent state can be used as an effective control variate for the score function gradient of the discrete latent states since the two are highly correlated. This allows us to obtain an unbiased gradient estimator. 

The score function gradient of the ELBO with respect to the continuous and discrete latent states are highly correlated when the continuous latent state is a good approximation of the discrete, as can be seen below.
\begin{equation*}
g_{SF}(z_d) = E_{q(z_d)}\left[L(x, z_d)\nabla_{\theta_1} log q(z_d; \theta_1)\right] = E_{q_c(z_c)}\left[L(x, H(z_c))\nabla_{\theta_1} log q_c(z_c; \theta_1)\right]
\end{equation*}

The score function gradient of the relaxed ELBO is:
\begin{equation*}
g_{SF}(z_c) = E_{q_c(z_c)}\left[L(x, z_c)\nabla_{\theta_1} log q_c(z_c; \theta_1)\right]
\end{equation*}
Where q$_c$ represents the probability density over the continuous latent states, $g_{SF}(z_d)$ represents the gradient of the ELBO with respect to $\theta_1$ with discrete latent states and $g_{SF}(z_c)$ with continuous latent states. This suggests that $g_{SF}(z_c)$ can be used as a control variate for 
g$_{SF}(z_d)$.
\\

The unbiased low variance gradient estimator can be obtained as:
\begin{equation*}
g = g_{SF}(z_d) - g_{SF}(z_c) + g_{REP}(z_c)
\end{equation*}

The term inside the expectation of $g_{SF}(z_d)$ is used by the Monte-Carlo samples to obtain gradient estimates, decreasing its magnitude decreases the variance among the gradient estimates. A control variate does just this without adding any bias to the overall gradient. In order for g$_{SF}(z_c)$ to be used to decrease the magnitude of the term inside the expectation term of $g_{SF}(z_d)$, it has to be expressed in terms of z$_d$, this can be done using conditional expectations as:
\begin{equation*}
g_{SF}(z_c) = E_{q(z_d)}\left[E_{q_c(z_c | z_d)}\left[L(x, z_c)\right]\nabla_{\theta_1} log q(z_d; \theta_1) + \nabla_{\theta_1}E_{q_c(z_c | z_d)}\left[L(x, z_c)\right]\right] 
\end{equation*}

Thus the unbiased gradient estimator can derived to be:
\begin{equation*}
\nabla_{\theta_1} ELBO = E_{q(z_d)}\left[\left(L(x, z_d) - L(x, z_c)\right)\nabla_{\theta_1} log q(z_d; \theta_1) - \nabla_{\theta_1}E_{q_c(z_c | z_d)}\left[L(x, z_c)\right]\right] 
\end{equation*}
\begin{equation*}
+ \nabla_{\theta_1}E_{q_c(z_c)}\left[L(x, z_c)\right]
\end{equation*}

The conditional distribution q$_c(z_c | z_d)$ and $q_c(z_c)$ allow reparameterization when the Gumbel softmax is used and hence their gradient estimates can be computed accurately. The expectation under q(z$_d$) is approximated using samples from z$_d$. Thus, this allows us to obtain an unbiased estimate of the gradient for learning. From the implementation perspective, the encoder and decoder architecture remain the same, the only change is in the ELBO being optimized. 
\\

Since the estimator is unbiased for any value of $\lambda$, the temperature parameter, $\lambda$ can be learnt by minimizing the variance among the gradient estimates. The variance of the gradient estimate can be evaluated using a single sample given as:
\begin{equation*}
Var(g) = 2g(\lambda) \nabla_{\lambda} g(\lambda)
\end{equation*}
Where g($\lambda$) is the gradient of the ELBO with respect to $\theta_1$) evaluated at $\lambda$. This makes $\lambda$ as a parameter instead of a user defined hyperparameter, making training easier. 

The closed form equation of the variance with respect to the $\lambda$ suggests that an arbitrarily complex function such as a neural network can be used either as a replacement or addition to the Gumbel Softmax, \cite{grathwohl2017backpropagation} shows results based on this idea. 

\chapter{Decreasing variance} \label{Decreasing variance}
Variational inference was made scalable by replacing all the expectation terms with Monte-Carlo approximations over the samples. The expected gradient in this method is estimated by averaging the gradient estimates of multiple samples obtained from the approximate posterior. Since the gradient of the ELBO has to be estimated in each step of the training process, we typically use a single/very few sample to keep training fast, however, using fewer samples results in a poorer approximation of the expectation. This leads to large variance in the gradient estimates for each step of learning. Such large fluctuating gradient estimates makes optimization hard. This chapter discusses techniques to reduce that variance. 
\\

The reparameterization trick notably exhibits the lowest variance compared to other unbiased gradient estimators, this small variance still however hampers learning. To reduce this variance, ELBO can be reformulated to expose the variance inducing term, which can be eliminated. More commonly, control variates are designed to combat this issue.  
\\

In order to train VAE, we are interested in learning the parameters $\theta_1$ and $\theta_2$ which correspond to the parameters of the encoder and decoder respectively. The gradient of ELBO with respect to $\theta_2$ can be estimated accurately using a single sample, since it does not depend on z, for this reason, we are interested in minimizing the variance in the gradient estimate of $\theta_1$. Using the same terminology, that $\phi$ represents the parameters of the approximate posterior, it is sufficient to find the gradient of the ELBO with respect to these parameters $\phi$, since this gradient can be easily backpropagated through the encoder network. 

The gradient of ELBO with respect to the parameters $\phi$ for a given sample z$_k$ is given as:
\begin{equation*}
\nabla_{\phi} ELBO_{z_k} = \nabla_{z_k} log P(x, z_k) \nabla_{\phi} z_k - \left[\nabla_{z_k} log q_{\phi}(z_k) \nabla_{\phi} z_k + \nabla_{\phi} log q_{\phi}(z_k)\right]
\end{equation*}

Assume that the approximate posterior was complex enough to capture the true posterior and the encoder learns to do so, in such a scenario, we want the gradient with respect to the parameters $\phi$ to be zero, since that setting of $\phi$ gives the best approximation, the true posterior itself. \cite{roeder2017sticking} notes that although this is the desired case, it is not what we obtain when we find the gradient of $\phi$ with respect to ELBO, instead we obtain the extra gradient of the posterior probability with respect to its parameters. this is easy to spot by setting the approximate posterior equal to the true posterior as:
\begin{equation*}
\nabla_{\phi} ELBO_{z_k} = log P(x) + \cancel{\nabla_{z_k} log P(z_k | x) \nabla_{\phi} z_k} - \cancel{\nabla_{z_k} log P_{\phi}(z_k | x) \nabla_{\phi} z_k} - \nabla_{\phi} log P_{\phi}(z_k | x)
\end{equation*} 
\\

The extra gradient term $\nabla_{\phi} log P_{\phi}(z_k | x)$ gives rise to incorrect gradient estimates. It is also worth noting that this is the score function gradient and the expected value of the score function gradient under the same distribution is zero [\ref{Score function derivative}], hence this term can be eliminated to reduce the incorrect estimate in the last parts of the learning process. In the initial stages, if it is highly correlated with the other terms, it can be retained so that it acts as a control variate to reduce the variance. 
\\

Other ways of reducing this variance includes using the first order Taylor approximation as the control variate, which is a method suggested in \cite{miller2017reducing}. However, this can be applied to only approximate posteriors with known Hessian or easy to compute Hessian with respect its parameters $\phi$. 
\newpage

\chapter{Improving the Objective}\label{improving_objective}
Ideally the marginal log probability over the observed data should be maximized to learn the probability distribution over the data, however, computing this is intractable, for this reason we maximize a lower bound on the true objective, the ELBO. The quality of learning results thus depend on the tightness of this lower bound and this bound becomes tighter as the KL divergence between the approximate posterior and the true posterior decreases. Chapter \ref{improving_encoder} discussed methods to make the approximate posterior complex enough to contain the true posterior, this chapter discusses ways to modify the objective to make the lower bound tighter, making it easier to learn the true posterior. 
\\

There are broadly two techniques to modify the objective and they are:
\begin{enumerate}
	\item Importance Sampling
	\item Monte Carlo Objectives
\end{enumerate}

\section{Importance Sampling}
ELBO is maximized by following the average gradient of the term $log \frac{P(x, z)}{q(z | x)}$ for samples z obtained using the encoder. When a particular sample z has a very low joint log probability log P(x, z), the distribution modelled by the encoder q(Z$|$X) is penalized strongly. However, even the perfect q(Z$|$X) can give rise to samples with low joint log probability, so the ELBO can never converge even when the approximate posterior is equal to the true posterior. The idea of importance sampling is to allow convergence and make training easier by placing importance on the samples such that bad samples from a good distribution q(Z$|$X) are penalized less. 
\\

The importance weighted objective is given as:
\begin{equation*}
ELBO_{IW} = E_{q(z_1, z_2, .. z_k | x)} \left[log \frac{1}{k} \sum_{i=1}^k \frac{p(x, z_i)}{q( z_i | x)}\right]
\end{equation*}
Here k refers to the number of samples and the expectation is under the probability distribution over the k different samples.
\\

Consider the weight for the i$^{th}$ sample, w(x, z$_i$) to be given by the ratio $\frac{p(x, z_i)}{q(z_i | x)}$, then the gradient of the ELBO with respect to the parameters $\theta$ is given as:
\begin{equation*}
\nabla_{\theta} ELBO_{IW} = E_{q(\epsilon_1, \epsilon_2, .. \epsilon_k | x)} \left[\nabla_{\theta} log \sum_{i=1}^k w(x, T(\phi, \epsilon)) \right]
\end{equation*}
\begin{equation} \label{importance weighted}
= E_{q(\epsilon_1, \epsilon_2, .. \epsilon_k | x)} \left[\frac{\sum_{i=1}^k w(x, T(\phi, \epsilon)\nabla_{\theta} log  w(x, T(\phi, \epsilon))}{\sum_{i=1}^k w(x, T(\phi, \epsilon)} \right]
\end{equation}

Here T($\phi$, $\epsilon$) refers to the reparameterization of the sample z in terms of the output of the encoder $\phi$ and a random sample $\epsilon$ obtained from the standard normal.
\\

From \eqref{importance weighted} it is easy to see how the magnitude of the gradient is scaled by the weight, which is low when the joint log probability is low, thus the encoder penalized only when the average estimate of the joint log probability is low, this makes training easier.
\\

Importance weighting can also be viewed as a transformation of the approximate posterior to a modified posterior that can learn a better model of the low probability regions of the true posterior, more details can be found in \cite{cremer2017reinterpreting}.

\section{Monte Carlo Objectives}
A Monte Carlo objective is the log of any unbiased lower bound estimator of the marginal log probability of the data. Based on this definition, ELBO is a Monte Carlo objective. Obtaining a generalized notion of the objective allows systematic study of the objectives to understand the effect of a particular objective on a particular objective.
\\

The work \cite{maddison2017filtering} uses Monte-Carlo objectives to prove that the first order Taylor approximation of the difference between the marginal log probability of the observed data and the Monte Carlo objective is proportional to the variance of the Monte-Carlo objective. Using this idea, it constructs filtering variational objectives for sequential data which are known to have lower variance than ELBO. A description of this method requires an understanding of particle filters and studying the relationship between the variance of various estimators which is beyond the scope of this report. 

\chapter{Conclusion}
The paper discussed how latent variable models can be used to learn both the probability distribution of the data and uncover hidden structure in it. Variational autoencoders was discussed as a method to learn latent variable models over the data, the  core research directions in the field was discussed as well. 

It should be noted that the report does not talk about applications of VAEs to various data problems, certain tricks have to be applied to get neural networks to work with such a variety of problems and \cite{Goodfellow-et-al-2016} is a better resource for such material. VAEs can be viewed from an information theoretic principle where the latent state can be seen as the minimum length encoding of the input, this view and its applications in information theory were also not discussed to keep the report more accessible. Please refer to\cite{zhao2017infovae} for more details. VAEs are also used in conjuction with Markov Chain Monte Carlo method which is another method to obtain samples from complex probability distributions, these methods were not discussed to prevent digression into  MCMC methods and fluid mechanics. More information can be found in \cite{salimans2015markov}, \cite{domke2017divergence}. To keep the material more accessible operator inference \cite{ranganath2016operator} another generalization of ELBO is not mentioned in the report.

In terms of practical problems, VAEs are hard to train and any research that improves optimization would be a valuable addition to the field, there are also no studies done on deciding the kind of neural network architectures and factors that go into choosing among the different techniques to improve the complexity of the base models. In terms of theory, it would be nice to quantify the quality of the approximation in terms of how far the approximation is from maximum likelihood estimation. 

We would like to emphasize that latent variable modelling is just one way to learn probability distributions of the data, recently implicit models \cite{tran2017hierarchical}, \cite{goodfellow2014generative} have gained a lot of popularity. They learn without maximum likelihood estimation. It would be interesting to establish theoretical relationships accross different methods of learning probability distributions. 

\chapter{Appendix}
\section{Score function derivatives} \label{Score function derivative}
The derivative of the ELBO with respect to the parameters of the encoder $\theta_1$ are given as:
\begin{equation*}
\nabla_{\theta_1} ELBO = \sum_{z} \left[log P(x, z) - log q(z; \theta_1) \right]\nabla_{\theta_1} q(z) + q(z) \nabla_{\theta_1} log q(z; \theta_1)
\end{equation*}
The second term is zero:
\begin{equation*}
\sum_{z} \cancel{ q(z)} \frac{\nabla_{\theta_1} q(z)}{\cancel{q(z)}} = \nabla_{\theta_1} \sum_{z} q(z; \theta_1) = \nabla_{\theta_1} 1 = 0 
\end{equation*}
Using the log derivative trick, we can write it as:
\begin{equation*}
= \sum_{z} q(z) \left[log P(x, z) - log q(z; \theta_1) \right]\nabla_{\theta_1} log q(z) 
\end{equation*}

\section{Control variates} \label{control variates}
Consider the problem of optimizing the expected value of a function of a random variable x. When the gradient of this expectation cannot be calculated in the closed form, samples of x can be used to obtain a Monte Carlo approximation of the gradient. High variance among the gradient estimates of each of the samples of x lead to inconsistencies in the approximations, this makes optimization hard. Control variates decrease this variance to allow optimization. They are constants or functions of x with known expectations under the distribution of x and strong correlation with the function being optimized. Consider f to be the function of interest and c the control variate, the objective can be expressed using the control variate as:
\begin{equation*}
E_{p(x)}[f(x)] = E_{p(x)}[f(x) - c] + E_{p(x)}[c]
\end{equation*}

\section{Score function gradients vs reparameterization gradients}
The key idea is to ignore the fact that the latent state is a function of the encoder's parameters and treat it as an input to the decoder instead, this allows us to train VAEs without the need for the reparameterization trick and hence permits the use of discrete latent states. 
\\

Let the parameters of the encoder be denoted by $\theta_1$ and the parameters of the decoder by $\theta_2$, the evidence lower bound can be expressed in terms of the expectation under the approximate posterior as follows:
\begin{equation*}
ELBO(\theta_1, \theta_2) = E_{q(z|x; \theta_1)}\left[log P(x|z; \theta_2) + log P(z) - log q(z | x; \theta_1)\right]
\end{equation*}
Since the prior logP(z) is specified by the practitioner, it has no parameters. The equation also ignores that z is function is a function of $\theta_1$ and instead treats it as an input variable. 
\\

For learning, we require the gradient of the ELBO with respect to the parameters $\theta_1$ and $\theta_2$.

The gradient with respect to $\theta_1$ can be derived as follows:
\begin{equation*}
\nabla_{\theta_1} ELBO = \sum_{z} \nabla_{\theta_1}\left[\left(log P(x|z; \theta_2) + log P(z) - log q(z | x; \theta_1)\right)q(z|x; \theta_1)\right]
\end{equation*}
For clarity we can express $f(z;\theta_2) = log P(x|z; \theta_2) + log P(z)$
and use product rule to get:
\begin{equation*}
\nabla_{\theta_1} ELBO = \sum_{z}\nabla_{\theta_1}\left[f(z;\theta_2) - log q(z | x; \theta_1)\right]q(z|x; \theta_1) + \nabla_{\theta_1} q(z|x; \theta_1) \left(f(z;\theta_2) - log q(z | x; \theta_1)\right)
\end{equation*}

The derivative of the first term is zero as shown below:
\begin{equation*}
\sum_{z}\nabla_{\theta_1}\left[f(z;\theta_2) - log q(z | x; \theta_1)\right]q(z|x; \theta_1) = -\sum_{z}\nabla_{\theta_1}log q(z | x; \theta_1)q(z|x; \theta_1)
\end{equation*}
\begin{equation*}
= -\sum_{z}\frac{\nabla_{\theta_1}q(z | x; \theta_1)}{\cancel{q(z | x; \theta_1)}}\cancel{q(z | x; \theta_1)} = -\nabla_{\theta_1}\sum_{z}q(z | x; \theta_1) = \nabla_{\theta_1} 1 = 0
\end{equation*}

By writing $\nabla_{\theta_1} q(z|x; \theta_1)$ = $\nabla_{\theta_1} log q(z|x; \theta_1) q(z|x; \theta_1)$ we can express the gradient as:
\begin{equation*}
\nabla_{\theta_1} ELBO = \sum_{z} -\nabla_{\theta_1}log q(z | x; \theta_1)q(z|x; \theta_1)\left(f(z;\theta_2) - log q(z | x; \theta_1)\right)
\end{equation*}

Thus, we can express the gradient of ELBO with respect to its parameters as:
\begin{equation*}
\nabla_{\theta_1}ELBO = E_{q(z|x; \theta_1)}\left[\nabla_{\theta_1}log q(z|x; \theta_1)\left[log P(x|z; \theta_2) + log P(z) - log q(z | x; \theta_1)\right]\right]
\end{equation*}
\begin{equation*}
\nabla_{\theta_2} ELBO = E_{q(z|x; \theta_1)}\left[\nabla_{\theta_2}log P(x|z; \theta_2)\right]
\end{equation*}
This gradient with respect to $\theta_1$ is called the score function gradient and is commonly used in reinforcement learning.

These gradients allow training, however, treating z as an input varibale instead of a function of the encoder's output has the adverse effect of giving high variance gradients of the ELBO with respect to the parameters of the encoder $\theta_1$, so it is common to couple this score function gradient with the variance reduction techniques using control variates. 
\\

To understand the source of variance, it is helpful to contrast the gradient with respect to $\theta_1$ with the gradient in the presence of the reparameterization trick. The reparameterization trick allows us to express z as z($\theta_1$) = T($\phi$; $\epsilon$) where $\phi$ is the output of the encoder that depends on $\theta_1$. ELBO treating z as a parameter can be written as:
\begin{equation*}
ELBO(\theta_1, \theta_2) = E_{q(z|x; \theta_1)}\left[log P(x|z; \theta_1, \theta_2) + log P(z; \theta_1) - log q(z | x; \theta_1)\right]
\end{equation*}

Using the same terminology we can represent f(z; $\theta_1, \theta_2$) = $log P(x|z; \theta_1, \theta_2) + log P(z; \theta_1)$, the gradient with respect to this function with respect to $\theta_1$ is no longer zero, but is given as:
\begin{equation*}
\nabla_{\theta_1}f(z; \theta_1, \theta_2) = \nabla_{z} f(z; \theta_1, \theta_2) \nabla_{\theta_1} z(\theta_1) 
\end{equation*}
$\nabla_{z} f(z; \theta_1, \theta_2)$ is just the gradient of the decoder with respect to its input(minus the prior p(z; $\theta_1$)), the reparameterization trick, allows gradients from the decoder to flow through the encoder. The gradient of the ELBO with respect to $\theta_2$ remains the same, with $\theta_1$ it is given as:
\begin{equation*}
\nabla_{\theta_1} ELBO = E_{q(z|x; \theta_1)}\left[\nabla_{z} f(z; \theta_1, \theta_2) \nabla_{\theta_1} z(\theta_1) + \nabla_{\theta_1}log q(z|x; \theta_1)\left[f(z; \theta_1, \theta_2) - log q(z | x; \theta_1)\right]\right]
\end{equation*}

When $\theta_1$ changes, the encoder outputs a different set of parameters $\phi$ as the parameters of the approximate posterior. z is sampled using the distribution modelled by these parameters. The score function gradient evaluates the second term of the above equation, $\nabla_{\theta_1}log q(z|x; \theta_1)\left[f(z; \theta_1, \theta_2) - log q(z | x; \theta_1)\right]$ with the sampled value of z. There is considerable variance among the samples of z which is given by the parameters $\phi$ of the distribution, however, since there is no way to relate z to $\phi$ in score function gradients, this variance cannot be captured, this leads to high variance in the gradients of the score function. 
\\

With the introduction of the reparameterization function, gradients of z explicitly flow through the parameters $\phi$ that produced it, thus even though the gradient is still evaluated on a sample z, the effect of evaluating on this particular sample is captured by relating the sample to the parameters. The only source of variance in this case is the noise distribution, which being a standard normal is quite small. For this reason, the reparameterization function obtains the lowest gradient estimate and allows easy training. 
\\

To decrease the variance of the score function we would have to find a way to relate the sample to the parameters, many approaches to train discrete latent states do that and will be discussed in the coming sections. Score function gradients on the other hand control the degree of variance using control variates. The most common approach to control the variance is through Rao-blackwellization. The use of control variates will not be discussed in detail here, more details about the method can be found in \cite{ranganath2014black}

\addcontentsline{toc}{chapter}{References}
\bibliographystyle{plain}
\bibliography{papers}
\nocite{*}
\end{document}